\documentclass[conference]{IEEEtran}
\IEEEoverridecommandlockouts
\usepackage{cite}
\usepackage{amsmath,amssymb,amsfonts}
\usepackage{algorithmic}
\usepackage{graphicx}
\usepackage{textcomp}
\usepackage{xcolor}
\usepackage{booktabs}
\usepackage{easyReview}
\usepackage{multirow}
\usepackage{graphicx}
\usepackage{url}
\usepackage{enumitem}
\usepackage{hyperref}

\def\BibTeX{{\rm B\kern-.05em{\sc i\kern-.025em b}\kern-.08em
    T\kern-.1667em\lower.7ex\hbox{E}\kern-.125emX}}
\begin{document}

\title{
Veterans' Mental Health Crisis Severity Annotation: Do LLMs Agree More Than Human Experts?
}

\makeatletter
\makeatother

\author{\IEEEauthorblockN{K M Sajjadul Islam}
\IEEEauthorblockA{Computer Science \\
Marquette University\\
sajjad.islam@marquette.edu}
\and
\IEEEauthorblockN{Pranav Yaramalla}
\IEEEauthorblockA{North Carolina School of \\
Science and Mathematics \\
pranu.yaramalla@gmail.com}
\and
\IEEEauthorblockN{Nadiyah Johnson}
\IEEEauthorblockA{Computer Science\\
Marquette University\\
nadiyah.johnson@marquette.edu}
\and
\IEEEauthorblockN{Mark Flower}
\IEEEauthorblockA{Veteran Peer Specialist \\
Captain John D Mason Peer Support Program\\
markflower999@gmail.com}
\and
\IEEEauthorblockN{Praveen Madiraju}
\IEEEauthorblockA{Computer Science \\
Marquette University\\
praveen.madiraju@marquette.edu}
\and
\IEEEauthorblockN{Zeno Franco}
\IEEEauthorblockA{Emergency Medicine \\
Medical College of Wisconsin\\
zfranco@mcw.edu}
}

\maketitle

\begin{abstract}
Large language models (LLMs) offer a scalable approach for annotating mental health and crisis-related social media content, but their reliability in safety-critical tasks remains unclear. This study evaluates LLM-based crisis-severity annotation of veteran-related social media posts against annotations from eight Veteran Subject Matter Experts (SMEs). Using 149 tweets and five severity categories, we evaluate 15 open-source and 10 proprietary LLMs across eight prompt variants, examining annotation performance, inter-model agreement, prompt sensitivity, and high-severity crisis detection. Human reviewers showed higher ordinal than nominal agreement ($\alpha=0.494$), indicating that disagreements often occurred between neighboring severity levels. In contrast, selected LLMs demonstrated substantially higher inter-model consistency, with ordinal agreement ($\alpha$) reaching $0.8336$ under the baseline prompt. Proprietary models generally showed better agreement with expert annotations and greater stability than open-source models. GPT-5.6-Terra achieved the best overall performance, while the open-source Qwen2.5-7B showed the strongest performance in identifying Acute Crisis with Suicidal Ideation cases. However, higher model consistency did not necessarily correspond to agreement with human judgments or reliable detection of rare high-severity cases. Label-flipped demonstration experiments also revealed model-specific sensitivity to misleading in-context examples, while explanatory demonstrations did not consistently improve performance. Overall, LLMs show promise as scalable complements to expert annotation, but require careful evaluation before use in high-stakes mental health applications.
\end{abstract}


\subsection*{Data and Code Availability}
\noindent Code: {\footnotesize
\url{https://github.com/sajjadIslam2619/LLM-Data-Annotation}}

\noindent VetCrisis Dataset: {\footnotesize \url{https://huggingface.co/datasets/SajjadIslam/VetCrisis}}

\section*{Introduction}
Social media platforms generate massive amounts of user-generated text that can be leveraged to study mental health and crisis-related behavior at population scale \cite{he2026mental}. However, manually annotating such data for mental health and crisis signals is expensive, time-consuming, and difficult to scale, especially when expert judgment is required. Crisis-severity annotation is particularly challenging because posts often convey distress indirectly, rely on context, and use ambiguous or linguistically subtle expressions such as sarcasm, slang, or emojis \cite{qadir2026triaging}. In this setting, annotation errors are especially consequential for high-severity crisis categories, where misidentifying critical self-harm or suicidal ideation may limit the effectiveness of downstream triage and intervention efforts.

Subject Matter Expert (SME) annotation is commonly used to establish reliable labels for mental-health and crisis-related content. However, such content is inherently subjective and strongly context-dependent, and different experts can interpret the same linguistic signal in divergent ways. Prior work in mental-health natural language processing (NLP) and affective computing has repeatedly observed only moderate inter-rater reliability among trained annotators, even under shared guidelines, and has shown that disagreement can reflect coherent but incompatible interpretive frameworks rather than simple annotation error. Under this view, annotator disagreement is not necessarily a failure of the annotation process; it can instead reveal genuine ambiguity and subjectivity in the underlying task \cite{arnaiz2026between}.

Large language models (LLMs) are increasingly used for text classification and annotation, including in health-related domains, where they can operate via prompting without task-specific fine-tuning \cite{islam2025llm}. In this paradigm, annotation codebooks or crisis taxonomies are incorporated into prompts, and the model assigns labels based on its pretrained knowledge, sometimes with a small number of examples. However, LLM behavior is sensitive to prompt formulation: changes in wording, framing, label order, or detail can shift outputs, a phenomenon described as prompt brittleness. Thus, LLMs are best viewed not as drop-in replacements for human annotators but as scalable complements whose outputs must be carefully characterized \cite{ccelebi2026much}.

Rather than asking whether LLMs are categorically ``better'' annotators than humans, we ask whether they can produce reliable and consistent crisis-severity annotations when human annotators themselves substantially disagree. We treat SME annotations as a multi-perspective signal and examine LLM behavior along several dimensions that may not align: agreement with the SME-derived reference labels, consistency across models, robustness to prompt and demonstration changes, and sensitivity to high-severity crisis cases. This distinction is important because aggregate annotation performance alone may not reveal how models behave when annotation categories are ambiguous, human reviewers disagree, or severe crisis cases are rare. Importantly, this study evaluates agreement with an SME-derived social-media annotation protocol and does not assess clinical diagnosis, imminent-risk assessment, or real-world intervention effectiveness.

In this study, we make four contributions:

\begin{enumerate}[leftmargin=20pt, label=(\arabic*), itemindent=0pt]
    \item Characterize disagreement among eight Veteran SMEs and construct SME-derived reference labels for crisis-severity annotation. (see section `Data')

    \item Systematically evaluate open-source and proprietary LLMs across multiple prompting strategies. (see section `Prompting')

    \item Compare human and LLM annotation behavior through inter-rater agreement, disagreement cases, and high-severity crisis detection. (see section `Results and Discussion')

    \item Examine model sensitivity to altered and explanatory few-shot demonstrations to characterize how in-context examples influence annotation behavior. (see section `LLM Behavior Under In-Context Demonstration Perturbations')
\end{enumerate}

\section*{Related Studies}

\textbf{Mental Health and Crisis Detection from Social Media.} Social media has emerged as a useful source for identifying mental-health and suicide-related concerns, particularly among populations that may be less likely to disclose risk in traditional healthcare settings. Prior studies have applied Transformer-based approaches to classify mental-health conditions and to distinguish self-harm and suicide-related expressions, with findings highlighting the importance of contextual information when interpreting ambiguous or hyperbolic language. \cite{qadir2026triaging} \cite{islam2025multimentalroberta} In a veteran-specific setting, Zuromski et al. investigated suicide-related content among U.S. servicemembers and veterans using posts from RallyPoint, a military-specific social-media platform. Their study further showed that contextual and metadata information can support the identification of suicidal thoughts and behaviors, while ambiguous expressions remain challenging to interpret. \cite{zuromski2024detecting} Together, these studies highlight the potential of computational language models for mental-health and crisis-related analysis while underscoring the complexity of interpreting nuanced social-media content.

\textbf{Human Annotation and Annotator Disagreement.} Human annotation remains important for mental-health and crisis-related social-media research because such content is often ambiguous, context-dependent, and requires domain expertise. Prior studies have shown that disagreement between annotators can reflect meaningful differences in interpretation rather than simple annotation errors, and that majority voting may obscure these perspectives. \cite{davani2022dealing,sandri2023don} Careful annotation scheme development, including expert involvement, iterative refinement, and validation, is therefore essential for producing reliable labels for complex mental-health constructs. \cite{chancellor2021suicide} These findings highlight the need to examine human disagreement when establishing ground truth, particularly for subjective and safety-critical annotation tasks.

\textbf{LLMs for Text Annotation and Mental Health Applications.} Recent studies have investigated LLMs as scalable alternatives to conventional human annotation for text classification. ChatGPT has demonstrated competitive performance with human annotators across diverse NLP tasks, while subsequent work has shown that LLMs can provide effective annotations in specialized settings \cite{gilardi2023chatgpt}. Open-source LLMs have similarly shown competitive performance with proprietary models, highlighting their potential for scalable annotation \cite{alizadeh2023open}. In mental-health applications, Settanni et al. compared multiple LLMs with trained clinicians for assessing the urgency of psychological distress in social-media posts, finding substantial model-specific differences and a tendency for LLMs to overestimate urgency relative to human judgments \cite{settanni2025assessing}. 

These findings demonstrate the potential of LLMs for mental-health annotation while highlighting the need to evaluate their agreement with domain experts, particularly for subjective and safety-critical crisis-severity tasks.

\textbf{Prompt Sensitivity and In-Context Learning.} In-context learning (ICL) enables LLMs to perform classification by using labeled demonstrations provided within the prompt, without updating model parameters. However, model behavior can be sensitive to how demonstrations and instructions are selected, organized, and presented. Prior studies have shown that the choice and ordering of in-context examples can substantially affect model predictions, while recent work further demonstrates that the position of demonstrations within a prompt can introduce prediction variability. \cite{liu2023towards,schoch2025good,cobbina2025show} Explanation-based ICL provides another way to influence model behavior by encouraging models to consider reasoning beyond superficial patterns. Honda and Oka showed that systematically exploring explanations for alternative labels can improve robustness, although their qualitative analysis also revealed that additional reasoning may produce plausible but unsupported inferences\cite{honda2025exploring}.

These findings highlight the importance of examining prompt configuration and reasoning behavior when evaluating LLMs as annotators, particularly for subjective classification tasks.

\section*{Methodology}

\subsection{Data}\label{sec:data}

\textbf{Data Collection.} To support the veteran crisis detection study, tweets were collected from Twitter using veteran-specific search queries developed through grounded theory analysis and expert knowledge of military terminology. Grounded theory analysis was used to identify population-specific concepts, experiences, and linguistic patterns associated with veteran crisis, which informed the development of veteran-specific search queries. These queries incorporated veteran self-identification phrases, crisis-specific markers, and military terminology such as service branches, bases, divisions, and common abbreviations. The collected tweets underwent text preprocessing, including conversion of HTML-encoded content to plain text, removal of user mentions and URLs. In addition, manually reviewed samples were assessed for emotional expression, veteran identification, authenticity, and crisis- or post-traumatic stress disorder (PTSD)-related content, with the resulting feedback used to iteratively refine the search queries. This process produced approximately 4,500 veteran-related tweets\cite{johnson2024crisisbert}. For the present study, a subset of approximately 149 tweets was selected. 

To assess the diversity of the selected dataset, we computed the Vendi Score, a diversity metric that quantifies the effective number of distinct instances based on their pairwise similarity \cite{friedman2022vendi}. The dataset obtained a Vendi Score of 125.6 out of 149, corresponding to approximately 84.3\% effective diversity, indicating substantial variation. Rather than removing noisy or near-duplicate tweets, we intentionally retained them to evaluate LLM robustness and consistency under realistic, imperfect data conditions. In particular, retaining near-duplicates allows us to examine whether models produce consistent labels for semantically similar inputs. An example of a near-duplicate pair and the corresponding LLM predictions is presented in Table~\ref{tab:llm-predictions-L5}. Reviewer identities and other personally identifiable information were removed to preserve privacy.

\begin{table}[tb]
\centering
\caption{Annotation Categories, Crisis Severity Levels, and Reference Label Distribution}
\label{tab:annotation_severity_distribution}
\begin{tabular}{p{5.5cm}cc}
\toprule
Annotation Category/Label & Severity & Count \\
\midrule
L1 - No Concern & 1 & 100 \\
L2 - Distressed Self-Disclosure & 2 & 19 \\
L3 - Self-Identified Crisis Symptoms and Seeking Support & 3 & 15 \\
L4 - Self-Identified Crisis Symptoms but Not Seeking Support & 4 & 0 \\
L5 - Acute Crisis with Suicidal Ideation & 5 & 6 \\
L6 - Not Relevant to the Research & -- & 9 \\
L7 - Other & -- & 0 \\
\midrule
Total & & 149 \\
\bottomrule
\end{tabular}
\flushleft
{
\textit{Note:} L1--L5 constitute the ordered crisis-severity taxonomy, with increasing severity from L1 to L5. L6 and L7 are categorical non-severity outcomes. The absence of L4 in the final reference labels reflects the documented difficulty of this category and its limited representation in prior annotation work \cite{johnson2024crisisbert}.}
\end{table}

\textbf{Ethical Considerations.} The study protocol was reviewed and determined exempt by the Marquette University Institutional Review Board (IRB) under secondary analysis of publicly available, de-identified data. User handles, hyperlinks, and identifiable markers were removed prior to analysis to safeguard participant anonymity.

\textbf{Taxonomy Development.} The seven-category crisis-severity taxonomy used in this study originated from prior work and was refined through two iterations using computational grounded theory and veteran peer-mentor feedback \cite{johnson2024crisisbert}. The refinement improved label clarity and interpretability; for example, ``Non-Alarming'' was revised to ``No Concern'', and ``Crisis Suicidal Final'' to ``Crisis Suicidal''. The prior study also identified challenges in distinguishing categories such as L4, as well as class overlap and limited representation that affected annotation agreement. The resulting taxonomy consists of five ordered crisis-severity levels (L1--L5) and two non-severity categories (L6--L7) (see Table \ref{tab:annotation_severity_distribution}), which serve as the labeling framework for this study.

\textbf{Expert Data Annotation.} The selected tweets were independently annotated by eight Veteran SMEs using a structured annotation protocol. The SME data labelers were invited to contribute their time and expertise based on their roles as veteran peer mentors. The group brought a range of lived and professional experience related to the challenges faced by U.S. military veterans, including experiences with mental-health crisis and non-crisis situations. This diversity of experience may have contributed to variability in individual ratings. Each annotator reviewed a tweet and assigned one of seven predefined categories, with the annotation categories and corresponding severity levels summarized in Table~\ref{tab:annotation_severity_distribution}. The severity levels 1--5 form an ordered continuum of increasing crisis severity, while the remaining two categories are treated as categorical non-severity outcomes. Annotators could also provide additional notes for ambiguous or exceptional cases. This multi-reviewer annotation process produced the reference labels used in this study.

\begin{table}[tb]
  \caption{Inter-Rater Reliability Before and After Label Consolidation}
  \label{irr-comparison-table}
  \centering
  \begin{tabular}{p{3.3cm}ccc}
    \toprule
    Metric & 7 Labels & 6 Labels & 5 Labels\\
    \midrule
    Majority Agreement ($\geq$5/8)  & 59.06\% & 61.74\% & 81.20\%\\
    Full agreement ($8/8$) & 0.00\% & 0.00\% & 32.88\%\\
    Fleiss' $\kappa$ & 0.1430 & 0.1442 & 0.2913\\
    Krippendorff's $\alpha$ (nominal) & 0.1437 & 0.1449 & 0.2919\\
    Krippendorff's $\alpha$ (ordinal) & \textbf{0.4941} & 0.4941 & 0.4467\\
    \bottomrule
  \end{tabular}
  \flushleft
  {
  \textit{Note:} The 6-label scheme consolidates L6 and L7 into a single label, while the 5-label scheme further consolidates L1, L6, and L7. Majority agreement requires at least 5 of 8 reviewers to select the same label. Fleiss' $\kappa$ measures inter-rater agreement among multiple reviewers beyond chance \cite{fleiss1971measuring}. Krippendorff's $\alpha$ measures inter-rater reliability and accommodates different measurement levels, including nominal and ordinal data \cite{krippendorff2004reliability}.}
\end{table}

\textbf{Inter-Rater Reliability.} As shown in Table~\ref{irr-comparison-table}, inter-rater reliability (IRR) measures the consistency of labeling decisions across independent annotators. For the original seven-label scheme, majority agreement was 59.06\%, with no tweets receiving unanimous agreement. Nominal Krippendorff's $\alpha$ indicates low agreement beyond chance. This level of agreement is consistent with challenges documented in the prior taxonomy-development work, which reported lower-than-expected inter-rater agreement and identified overlapping class attributes and limited representation of some categories as contributing factors \cite{johnson2024crisisbert}. Thus, the majority-vote labels used in this study should be interpreted as aggregate human judgments rather than unambiguous ground truth.

To examine the extent to which category structure contributed to disagreement, we conducted sensitivity analyses using six- and five-label schemes. Consolidating L6 (Not Relevant to the Research) and L7 (Other) produced only a negligible change in nominal agreement. Further consolidation of L1, L6, and L7 increased nominal agreement more substantially. These results suggest that some disagreement was associated with distinctions among the non-severity categories, although substantial categorical disagreement remained. We therefore retained the original seven-label taxonomy for the primary analysis because it preserves the distinctions defined by the veteran-informed annotation framework.

\begin{figure*}[tb]
\centerline{\includegraphics[width=0.9\textwidth]{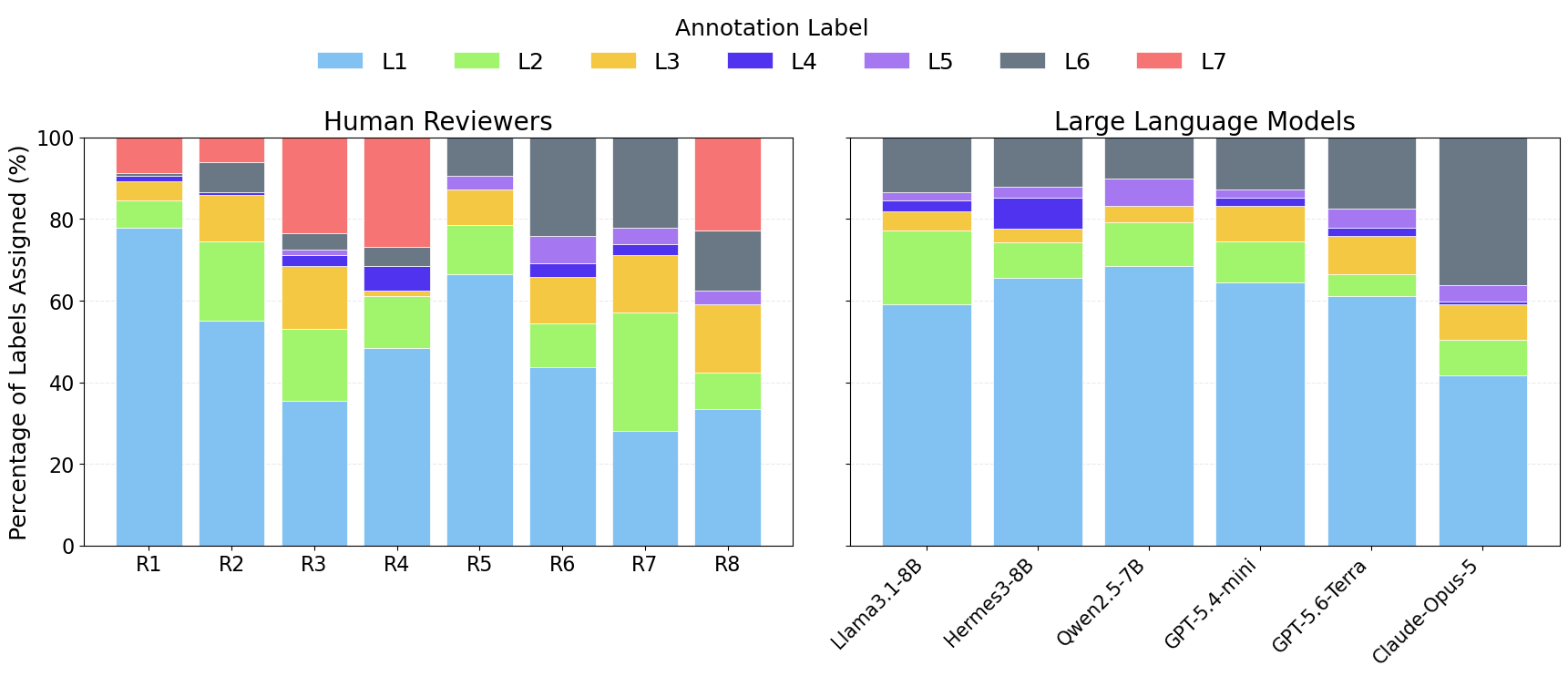}}
\caption{Label distributions of the eight Veteran SME reviewers and six selected LLMs under Prompt G. The stacked percentages show the relative frequency with which each reviewer/LLM assigned the seven annotation categories.}
\label{fig:reviewer-distribution}
\end{figure*}

Because L1--L5 represent an ordered progression of crisis severity, we additionally evaluated ordinal agreement using Krippendorff's $\alpha$. As shown in Fig.~\ref{fig:reviewer-distribution}, the reviewers exhibited noticeable differences in their label distributions, particularly across the severity levels. The ordinal $\alpha$ of 0.4941 was higher than the nominal value, indicating that disagreements were often between adjacent or nearby severity levels rather than across distant levels. Thus, although exact categorical agreement was limited, the reviewers showed meaningful consistency in their assessment of relative crisis severity.

\textbf{Reference Label Construction by Majority Vote.} To obtain a single reference label for each tweet, we used majority voting across the eight human reviewers. The reference label represents the label receiving the largest number of reviewer votes; thus, it aggregates multiple human judgments into one label for comparison with LLM annotations. Of the 149 tweets, 140 (93.96\%) had a unique majority-vote outcome, while 9 (6.04\%) resulted in ties. This differs from the majority-agreement measure in Table~\ref{irr-comparison-table}, which requires at least five of eight reviewers to assign the same label and is used to quantify inter-rater agreement. Majority voting, in contrast, is a label-construction procedure that provides one reference label even when reviewers do not reach strong consensus. Therefore, the resulting reference labels should be interpreted as aggregate human judgments rather than unambiguous ground truth.

Among the tied cases, 7 were two-way ties and 2 were three-way ties. Ties were resolved by selecting the more severe label among the tied categories, following a conservative strategy for crisis-severity annotation. For example, the tweet ``I'm done \#bipolar \#bipolarclub'' received a 2--2 split between L2 (Distressed Self-Disclosure) and L5 (Acute Crisis with Suicidal Ideation), as shown in Table~\ref{tab:tied-votes-selected}. Because L5 represents greater severity, it was selected as the reference label. This procedure ensures that every tweet receives a single reference label while preserving a conservative approach to ambiguous crisis cases.

The resulting reference distribution is highly skewed toward L1 (No Concern), with 100 of the 149 tweets assigned to this category (see Table \ref{tab:annotation_severity_distribution}). Such class imbalance is common in social media studies of crisis and suicide-related content, where posts expressing acute risk or crisis are relatively rare compared with non-crisis content \cite{zuromski2024detecting}. 

\subsection{Prompting}\label{Prompting}
Prompting provides instructions and contextual information that guide an LLM toward a desired task and output. It offers a practical way to leverage the capabilities of pretrained LLMs for a wide range of tasks without requiring task-specific training or fine-tuning \cite{alammar2024hands}. Prompts can specify the task, define target categories, provide examples, encourage reasoning, and constrain the expected output format. These components can reduce ambiguity and help models interpret task requirements more consistently. However, the choice and combination of prompt components can substantially influence model behavior and performance. Therefore, we systematically vary prompt components to examine how different forms of guidance affect crisis-severity classification.

We evaluated the eight prompt variants (A--H) across 15 popular open-source models deployed locally through the Ollama library\footnote{https://ollama.com/library}. We additionally evaluated all prompt variants using proprietary models accessed through APIs, including eight GPT model variants through the OpenAI API\footnote{https://developers.openai.com/api/docs/models} and two Claude model variants through the Anthropic API\footnote{https://platform.claude.com/docs/en/models/overview}. The complete set of model--prompt combinations and their corresponding Micro-F1 scores are reported in Table~\ref{tab:f1-prompt}. This setup allows us to examine the effects of prompt variations across both locally deployed open-source models and proprietary models.

\textbf{Prompt A -- Reviewer-Matched. (Baseline Prompt)} Provides the models with the same category information available to the human reviewers, using the short label definitions but no worked examples or explicit reasoning instructions. This prompt establishes the most direct human--model comparison by aligning the information provided to both groups while allowing the models to make their classifications independently.

\textbf{Prompt B -- Label-Only.} Provides only the list of label names, without definitions, examples, or reasoning guidance. This serves as the baseline for measuring how effectively models can infer the intended distinctions among crisis-severity categories from the label vocabulary alone.

\textbf{Prompt C -- Short-Definition.} Extends Prompt B by adding a concise, one-sentence definition for each label. This variant evaluates whether providing minimal semantic guidance improves classification over relying solely on the label names while keeping the prompt relatively compact.

\textbf{Prompt D -- Long-Definition.} Replaces the short definitions in Prompt C with detailed, paragraph-length descriptions of each category. By keeping examples and reasoning instructions absent, this prompt isolates the effect of providing more extensive category-specific information and examines whether additional definitional detail improves classification.

\textbf{Prompt E -- Few-Shot.} Builds on Prompt D by adding three worked examples, each consisting of a tweet paired with its corresponding label. The examples were selected to have relatively high reviewer agreement and were not drawn from the evaluation set. This prompt evaluates whether concrete demonstrations of the labeling task provide additional benefits beyond detailed category definitions.

\textbf{Prompt F -- Reasoning-Enhanced.} Extends Prompt E by instructing the model to briefly reason about the tweet's tone, intent, and relevant context before producing its classification. This variant examines whether an explicit reasoning scaffold improves the model's ability to apply the category definitions and few-shot demonstrations, particularly when distinguishing between neighboring crisis categories requires contextual interpretation.

\textbf{Prompt G -- Refined-Definition.} Revises the detailed definitions in Prompt D based on failure patterns identified during the earlier ablation experiments. In particular, the definitions for Labels 1 (No Concern), 3 (Self-Identified Crisis Symptoms and Seeking Support), and 5 (Acute Crisis with Suicidal Ideation) were refined to address recurring classification errors. Three new synthetic examples were also introduced to reinforce these distinctions. This prompt evaluates whether targeted, error-informed prompt refinement can improve classification beyond the original definitions.

\textbf{Prompt H -- Structured-Reasoning.} Uses the detailed definitions and three worked examples from Prompt E and additionally requires the model to provide brief reasoning followed by a structured final label. Unlike Prompt F, which extends the few-shot prompt with a reasoning instruction, this variant explicitly enforces both the reasoning step and a fixed output format. It therefore examines whether structured reasoning and output constraints affect classification while keeping the underlying definitions and demonstrations unchanged.

\begin{table*}[tb]
\caption{Micro-F1 Across Prompt Variants}
\label{tab:f1-prompt}
\centering
\begin{tabular}{c l cccccccccccc}
\toprule
Type & Model / Prompt & \textbf{A} & B & \textbf{C} & D & E & F & \textbf{G} & H 
& Mean & SD & Avg. Rank \\
\midrule

\multirow{18}{*}{\rotatebox[origin=c]{90}{\textit{Open-source local models (Ollama)}}}

& aya-expanse-8b 
& 0.714 & 0.604 & -- & 0.468 & 0.746 & -- & 0.515 & 0.736 
& 0.630 & 0.120 & 6.000 \\

& command-r7b-7b 
& 0.581 & 0.376 & 0.581 & 0.192 & 0.613 & 0.657 & 0.719 & 0.595 
& 0.539 & 0.171 & 8.938 \\

& deepseek-r1-8b 
& 0.298 & 0.190 & 0.298 & 0.375 & 0.344 & 0.406 & 0.742 & 0.486 
& 0.392 & 0.166 & 10.500 \\

& gemma3-4b 
& 0.479 & 0.423 & 0.479 & 0.304 & 0.753 & 0.486 & 0.763 & 0.379 
& 0.508 & 0.166 & 9.000 \\

& granite3.3-8b 
& 0.471 & 0.357 & 0.471 & 0.602 & 0.371 & 0.679 & 0.693 & 0.282 
& 0.491 & 0.161 & 10.625 \\

& \textbf{hermes3-8b} 
& 0.629 & 0.636 & 0.629 & 0.797 & 0.780 & 0.719 & 0.785 & 0.402 
& \textbf{0.672} & 0.131 & \textbf{4.375} \\

& internlm2-7b 
& 0.757 & 0.451 & -- & 0.523 & 0.707 & 0.351 & 0.803 & 0.425 
& 0.574 & 0.179 & 6.286 \\

& llama3-8b 
& 0.693 & 0.552 & 0.693 & 0.729 & 0.700 & 0.626 & 0.712 & 0.462 
& 0.646 & 0.094 & 6.250 \\

& \textbf{llama3.1-8b} 
& 0.721 & 0.586 & 0.721 & 0.764 & \textbf{0.824} & 0.781 & 0.777 & 0.632 
& \textbf{0.726} & 0.081 & \textbf{2.375} \\

& mistral-7b 
& 0.592 & 0.380 & 0.592 & 0.583 & 0.517 & 0.682 & 0.727 & 0.410 
& 0.561 & 0.121 & 8.375 \\

& olmo2-7b 
& 0.451 & 0.438 & 0.451 & 0.073 & 0.614 & 0.746 & 0.574 & 0.447 
& 0.474 & 0.195 & 9.625 \\

& phi4-mini-3.8b 
& -- & 0.370 & 0.335 & 0.403 & 0.459 & 0.213 & 0.258 & 0.486 
& 0.361 & 0.100 & 11.143 \\

& \textbf{qwen2.5-7b} 
& 0.729 & 0.479 & 0.729 & 0.757 & 0.757 & 0.657 & 0.757 & 0.458 
& \textbf{0.665} & 0.126 & \textbf{4.438} \\

& qwen3-8b 
& 0.649 & 0.423 & 0.649 & 0.613 & 0.631 & 0.704 & 0.791 & 0.373 
& 0.624 & 0.094 & 6.625 \\

& yi-6b 
& 0.727 & 0.284 & -- & 0.438 & -- & -- & 0.533 & 0.479 
& 0.493 & 0.161 & 9.200 \\

\cmidrule{2-13}

& Mean 
& 0.588 & 0.437 & 0.588 & 0.508 & 0.619 & 0.603 & \textbf{0.677} & 0.470 
& -- & -- & -- \\

& SD 
& 0.143 & 0.122 & 0.145 & 0.145 & 0.136 & 0.173 & 0.148 & 0.112 
& -- & -- & -- \\

\midrule

\multirow{13}{*}{\rotatebox[origin=c]{90}{\textit{Proprietary models (API)}}}

& gpt-3.5-turbo 
& 0.706 & 0.487 & 0.708 & 0.784 & 0.801 & 0.781 & 0.827 & 0.767 
& 0.733 & 0.108 & 4.625 \\

& gpt-4o 
& 0.706 & 0.476 & 0.699 & 0.719 & 0.787 & 0.807 & 0.786 & 0.810 
& 0.724 & 0.110 & 4.875 \\

& gpt-4o-mini 
& 0.669 & 0.500 & 0.648 & 0.725 & 0.734 & 0.761 & 0.731 & 0.770 
& 0.692 & 0.088 & 6.875 \\

& gpt-5.4 
& 0.771 & 0.542 & 0.711 & 0.596 & 0.756 & 0.713 & 0.531 & 0.715 
& 0.667 & 0.096 & 6.500 \\

& \textbf{gpt-5.4-mini} 
& 0.662 & 0.742 & 0.684 & 0.809 & 0.794 & 0.765 & \textbf{0.839} & 0.783 
& \textbf{0.760} & 0.061 & \textbf{4.188} \\

& gpt-5.6-luna 
& 0.733 & 0.630 & 0.730 & 0.504 & 0.671 & 0.549 & 0.760 & 0.540 
& 0.639 & 0.099 & 6.813 \\

& gpt-5.6-sol 
& 0.787 & 0.774 & 0.771 & 0.645 & 0.754 & 0.739 & 0.760 & 0.708 
& \textbf{0.742} & 0.046 & 4.813 \\

& \textbf{gpt-5.6-terra} 
& 0.748 & 0.769 & 0.751 & -- & 0.816 & 0.794 & 0.820 & 0.783 
& \textbf{0.783} & 0.029 & \textbf{2.500} \\

& \textbf{claude-opus-5} 
& 0.715 & 0.732 & 0.700 & 0.744 & -- & 0.817 & 0.636 & 0.814 
& 0.737 & 0.064 & \textbf{4.143} \\

& claude-sonnet-5 
& 0.607 & 0.288 & 0.679 & 0.679 & -- & 0.760 & 0.693 & 0.773 
& 0.640 & 0.165 & 7.857 \\

\cmidrule{2-13}

& Mean 
& 0.710 & 0.594 & 0.708 & 0.689 & 0.764 & 0.749 & 0.738 & 0.746 
& -- & -- & -- \\

& SD 
& 0.047 & 0.162 & 0.080 & 0.096 & 0.046 & 0.077 & 0.096 & 0.080 
& -- & -- & -- \\

\bottomrule
\end{tabular}

\vspace{3pt}
\parbox{\textwidth}{
\textit{Note}: A--H denote the eight prompt variants. Values are Micro-F1. The Mean column reports the mean Micro-F1 across available prompt variants for each model, with missing values excluded. SD denotes the sample standard deviation across available prompt variants. Average Rank denotes the mean of the model's ranks across available prompt variants, where rank 1 indicates the highest Micro-F1. For the prompt-level summary rows, Mean and SD are calculated across models for each prompt. ``--'' indicates no usable result due to a missing file or more than 50\% unparseable outputs. Bold values indicate the highest Micro-F1 within each model group for each prompt variant.}
\end{table*}

\subsection{LLM Configuration.} We evaluated both proprietary and locally deployed LLMs using the same task-specific annotation framework and prompt variants. OpenAI legacy models and Ollama-based models were queried with temperature set to 0, while GPT-5.x models used temperature 0 when supported by the API. Anthropic models were evaluated using the provider default temperature. No sampling seed was specified, and each model--prompt--tweet combination was evaluated once without repeated sampling or self-consistency aggregation. Local models were deployed through Ollama on two computing environments: a Windows-based system with an Intel Core i7 processor, NVIDIA RTX 2000 GPU, and 32 GB RAM, and a MacBook Air with approximately 16 GB of unified memory. Model-specific quantization and runtime checksums were not recorded. Additional implementation details, including model configurations, prompts, inference scripts, and experimental settings, are provided in the accompanying GitHub repository. 

\section*{Results and Discussion}\label{sec:Result-And-Discussion}

\begin{table}[tb]
\caption{Macro-F1 Scores of the Six Selected LLMs Under Prompts A and G}
\label{tab:macro-f1-selected}
\centering
\begin{tabular}{llcc}
\toprule
Type & Model/Prompt & A & G \\
\midrule
\multirow{3}{*}{Open-Source LLM}
    & hermes3-8b   & 0.547 & 0.611 \\
    & \textbf{llama3.1-8b} &  0.555 & 0.654 \\
    & qwen2.5-7B     & 0.542 & 0.569 \\
\midrule
\multirow{3}{*}{Proprietary LLM}
    & \textbf{gpt-5.4-mini}   & 0.584 & 0.750 \\
    & gpt-5.6-terra  &  0.517 & 0.699 \\
    & claude-opus-5 & 0.669 & 0.640 \\
\bottomrule
\end{tabular}
\end{table}

We evaluate the alignment of LLM-generated annotations with SME-derived reference labels using Micro-F1, Average Rank, and standard deviation (SD). Micro-F1 is the primary evaluation metric and provides an overall measure of alignment between model-generated and SME-derived labels across all tweets. This aggregate measure is appropriate because L4 and L7 have no instances in the reference labels, making class-specific evaluation of these categories uninformative. Average Rank captures the relative alignment of each model across prompt variants, with lower values indicating stronger performance, while SD measures variation in Micro-F1 across prompts, with lower values indicating greater stability. Table~\ref{tab:f1-prompt} summarizes these measures across all evaluated models. As a complementary analysis, Macro-F1 is also reported for the six selected models under Prompts A and G in Table~\ref{tab:macro-f1-selected}.

\textbf{Overall Model Performance.} Proprietary models generally show stronger alignment with the SME-derived reference labels than open-source local models. Across the eight prompts, the proprietary models achieve higher prompt-level mean Micro-F1 than the open-source models for every prompt, with the largest difference observed for Prompt E (0.764 vs.\ 0.619). Proprietary models also generally exhibit lower variability across prompt variants. At the individual-model level, GPT-5.6-Terra achieves the highest mean Micro-F1 (0.783) and the lowest SD (0.029) among all evaluated models, indicating both strong overall alignment and comparatively stable behavior across prompt formulations. The Macro-F1 results for the six selected models show a consistent overall pattern: proprietary models achieve higher Macro-F1 than the open-source models under both Prompt A and Prompt G, although individual model rankings differ between the two prompts. These results indicate that the observed advantage of proprietary models is not attributable solely to the dominant L1 category, but is also reflected in the more balanced Macro-F1 measure.

\textbf{Prompt Selection.} We select Prompts A and G for the subsequent analysis. Prompt A serves as the baseline because it provides the models with the same category information available to the human reviewers, establishing the most direct human--model comparison. Prompt G is selected as the enhanced prompt because it achieves the highest mean Micro F1 among the open-source local models (0.677) and also yields strong performance for the selected proprietary models, including GPT-5.4-mini (0.839) and GPT-5.6-Terra (0.820). Although proprietary models show relatively similar performance across Prompts E--H, with Prompt E achieving the highest mean Micro F1 (0.764), the differences among these prompts are modest. We therefore use Prompt G as the enhanced prompt for both open-source and proprietary models, enabling a common comparison between the baseline and the error-informed prompt while capturing the strongest prompt-level performance for the open-source models.

\textbf{Model Selection.} We select six models for detailed analysis based primarily on Average Rank, which captures consistency of relative alignment across prompt variants rather than performance under a single prompt. Among the open-source models, Llama3.1-8B achieves the best Average Rank (2.375), followed by Hermes3-8B and Qwen2.5-7B. Among the proprietary models, GPT-5.6-Terra achieves the best Average Rank (2.500), followed by Claude-Opus-5 and GPT-5.4-mini. These six models are therefore selected for the subsequent analyses of inter-model agreement, disagreement with SME reference labels, high-severity cases, and sensitivity to in-context demonstrations.

\begin{table}[tb]
\caption{Inter-model Reliability of the Six Selected LLMs Under Prompts A and G}
\label{tab:reliability-selected}
\centering
\begin{tabular}{l r rr}
\toprule
Metric & Prompt A & \multicolumn{2}{c}{Prompt G} \\
\cmidrule(lr){3-4}
& 6 Labels & 6 Labels & 5 Labels \\
\midrule
Majority agreement ($\geq$4/6)     
    & 86.58\% & 87.25\% & 91.95\% \\
Full agreement ($6/6$)              
    & 63.09\% & 51.68\% & 73.83\% \\
Fleiss' $\kappa$                    
    & 0.6619 & 0.6038 & 0.6490 \\
Krippendorff's $\alpha$ (nominal)   
    & 0.6619 & 0.6083 & 0.6498 \\
Krippendorff's $\alpha$ (ordinal)   
    & 0.8336 & 0.8533 & 0.8321 \\
\bottomrule
\end{tabular}

\parbox{\columnwidth}{
\vspace{2pt}
\textit{Note:} The six selected models are Llama3.1-8B, Hermes3-8B, Qwen2.5-7B, GPT-5.6-Terra, Claude-Opus-5, and GPT-5.4-mini. The models are treated as independent raters, and reliability is computed across their predictions for the evaluation tweets. For comparison with the LLM outputs, the human annotations use a 6-label scheme that consolidates L6 and L7; the LLM outputs contain no L7 assignments and therefore comprise six observed labels (L1--L6). The 5-label scheme additionally consolidates L1 with L6 and L7.}
\end{table}

\textbf{Inter-rater Consistency (Humans vs LLMs).} Fig.~\ref{fig:reviewer-distribution} compares the label distributions of the eight human reviewers and six selected LLMs. LLM annotations are less diverse than human annotations, with models concentrating their predictions on a smaller set of labels. Notably, none of the selected LLMs assigned any tweet to L7 (Other), whereas several human reviewers used this category. Overall, the figure suggests that LLMs exhibit more concentrated labeling patterns than human reviewers.

\begin{figure}[tb]
\centerline{\includegraphics[width=0.50\textwidth]{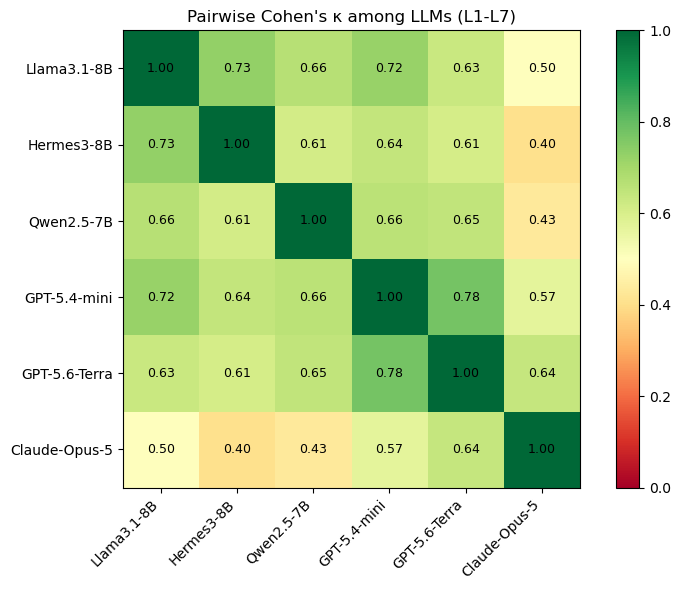}}
\caption{Pairwise Cohen's $\kappa$ Agreement Among LLMs Under Prompt G. Cohen's $\kappa$ measures pairwise inter-rater agreement beyond chance~\cite{cohen1960coefficient}.}
\label{fig:pairwise-cohen-k}
\end{figure}

\begin{table}[tb]
\caption{Inter-model Reliability of Open-Source and Hosted LLMs}
\label{tab:reliability-llm-comparison}
\centering
\begin{tabular}{lrr}
\toprule
Metric & Open-Source LLM & Proprietary LLM \\
\midrule
Majority agreement ($\geq$2/3)     & 97.99\% & 97.32\% \\
Full agreement ($3/3$)              & 73.83\% & 68.46\% \\
Fleiss' $\kappa$                    & 0.6676 & 0.6490 \\
Krippendorff's $\alpha$ (nominal)   & 0.6716 & 0.6498 \\
Krippendorff's $\alpha$ (ordinal)   & 0.8113 & 0.9297 \\

\bottomrule
\end{tabular}
\parbox{\columnwidth}{
\vspace{2pt}
\textit{Note:} The open-source LLM group consists of Llama3.1-8B, Hermes3-8B, and Qwen2.5-7B, while the proprietary LLM group consists of GPT-5.6-Terra, Claude-Opus-5, and GPT-5.4-mini. The models are treated as independent raters, and reliability is computed across their predictions under Prompt G.}
\end{table}

\begin{table*}[tb]
\caption{Tweets with Tied Human Reviewer Votes and LLM Votes from the 6 Selected Models}
\label{tab:tied-votes-selected}
\centering
\begin{tabular}{p{5cm}p{2.5cm}p{1.9cm}p{3cm}p{3cm}}
\toprule
Tweet &
Human Votes  &
Reference Label &
\multicolumn{2}{c}{LLM Votes} \\
\cmidrule(lr){4-5}
& (Tied) & (Tie Resolution) & Prompt A & Prompt G \\
\midrule

I'm done \#bipolar \#bipolarclub113 &
L2; L5 [2 votes] &
L5 &
L4; L5 [3 votes] &
L2; L4 [3 votes] \\

I meant I'm doing okay. \#PTSDChat &
L1; L2 [4 votes] &
L2 &
L1 (6 votes) &
L1 (6 votes) \\

Hi everyone \#PTSDChat113 &
L1; L3 [3 votes] &
L3 &
L1 (6 votes) &
L1 (6 votes) \\

My current mood is.....probably not safe enough to tweet about11 &
L2; L3; L5 [2 votes] &
L5 &
L2 (5 votes); L5 (1 vote)  &
L2 (3 votes); L4 (2 votes); L5 (1 vote) \\

Fuckin ideation won't go away HELP @BipolarClubRx \#bipolar111 &
L1; L3; L5 [2 votes] &
L5 &
L3; L5 [3 votes] &
L3 (4 votes); L5 (2 vote) \\

\bottomrule
\end{tabular}

\parbox{\textwidth}{
\vspace{2pt}
\textit{Note:} L1--L5 correspond to the five crisis-severity categories defined in Table~\ref{tab:annotation_severity_distribution}. Human votes are reported for the tied categories, with the number of votes shown in square brackets indicating that the displayed categories received the same number of votes. Reference Label is determined by selecting the higher-severity label among the tied human votes. LLM votes show the number of selected models predicting each label; parentheses indicate the vote count for each label.}
\end{table*}

Table~\ref{tab:reliability-selected} compares inter-rater agreement among the six selected LLMs with the human-reviewer agreement reported in Table~\ref{irr-comparison-table}. Because none of the LLMs assigned L7, LLM agreement is computed over the six observed categories (L1--L6), while the human agreement is reported over the full seven-category. The LLMs exhibit substantially higher nominal and ordinal agreement than the human reviewers. More importantly, the higher ordinal agreement for both groups indicates that disagreements often occur between neighboring severity levels rather than across distant levels. This pattern is particularly pronounced for the LLMs, suggesting that models are generally consistent in their assessment of relative crisis severity even when they differ in exact categorical assignments. However, greater inter-model agreement should not be interpreted as evidence of better alignment with the SME-derived reference labels. Inter-model agreement measures consistency among models, whereas Micro-F1 measures alignment with the same reference labels (see Table~\ref{tab:f1-prompt}).

The consolidation analysis further highlights the role of ambiguous non-severity categories. For both human reviewers and LLMs, consolidating L1, L6, and L7 improves categorical agreement, indicating that distinctions among these categories contribute to annotation disagreement. Notably, none of the selected LLMs assigned any tweet to L7 (Other), whereas human reviewers used this category, consistent with the less diverse LLM label distributions shown in Fig.~\ref{fig:reviewer-distribution}. Overall, LLMs demonstrate greater inter-model consistency than human reviewers, while disagreement remains concentrated around severity boundaries and ambiguous category distinctions.

\textbf{Inter-model Reliability (Open-Source vs. Proprietary LLM).} As shown in Fig.~\ref{fig:pairwise-cohen-k}, the selected LLMs generally exhibit moderate-to-high pairwise agreement under Prompt G, although agreement varies across model pairs. Notably, Claude-Opus-5 shows lower agreement with the other selected models. Table~\ref{tab:reliability-llm-comparison} shows similar nominal agreement between open-source and proprietary LLMs, while proprietary models exhibit substantially higher ordinal agreement. This suggests that the two groups are similarly consistent in their categorical annotations, but proprietary models show greater consistency when the ordered severity levels are considered.

\textbf{Disagreement Analysis (Humans vs. LLMs).} Table~\ref{tab:tied-votes-selected} presents the cases in which the human reviewers did not reach a clear majority and compares the resulting reference labels with predictions from the six selected LLMs under Prompts A and G. These cases are particularly challenging because the tweets contain ambiguous, contextual, or subtle signals of distress. We therefore use them to examine how LLMs behave when human annotations are themselves conflicting.

In the first example, ``I'm done \#bipolar \#bipolarclub113,'' the human reviewers were evenly split between L2 and L5, resulting in L5 after severity-based tie resolution. The LLMs also showed disagreement: under Prompt A, L4 and L5 received equal votes, while Prompt G produced a tie between L2 and L4. Although Prompt A included the reference label L5 among the model predictions, neither prompt produced a clear consensus on the higher-severity interpretation. This illustrates the difficulty of assigning an appropriate severity level to ambiguous expressions such as ``I'm done.''

The second and third examples illustrate cases in which the models consistently favored the lower-severity interpretation. For ``I meant I'm doing okay. \#PTSDChat,'' reviewers were split between L1 and L2, with L2 selected through tie resolution, whereas all six models selected L1 under both prompts. Similarly, for ``Hi everyone \#PTSDChat,'' reviewers were split between L1 and L3, resulting in L3 after tie resolution, while all six models selected L1 under both prompts. In the third case, participation in a PTSD-focused forum such as \#PTSDChat can itself provide an important contextual signal that the individual may be seeking attention, support, or engagement around their experiences. Therefore, the seemingly neutral phrase ``Hi everyone'' should not necessarily be interpreted in isolation, and L3 may be the more appropriate label when this contextual information is considered. These examples suggest that the models may place greater emphasis on the explicit wording of the tweets while underweighting contextual signals that influenced the human reviewers' severity judgments.

The fourth and fifth examples further highlight differences in recognizing severity cues. For ``My current mood is.....probably not safe enough to tweet about11,'' human reviewers selected L5 through severity-based tie resolution among L2, L3, and L5. However, the LLMs predominantly selected L2 under both prompts, with Prompt A producing five L2 votes and Prompt G producing three L2 votes. In contrast, the fifth example, ``Fuckin ideation won't go away HELP \\@BipolarClubRx \#bipolar111,'' was also resolved to L5 by the human reviewers, and Prompt G produced L5 for two of six models while the remaining four selected L3. These cases suggest that LLMs may recognize explicit indicators of distress but can still differ from human reviewers in assigning the appropriate severity level, particularly when the distinction depends on contextual or nuanced cues.

\begin{table*}[tb]
\caption{LLM Predictions for Tweets with Human Reference Label L5}
\label{tab:llm-predictions-L5}
\centering
\begin{tabular}{p{5.5cm}cccccc}
\toprule
Tweet &
{llama3.1-8b} &
{hermes3-8b} &
{\textbf{qwen2.5-7b}} &
{gpt-5.4-mini} &
{claude-Opus-5} &
{\textbf{gpt-5.6-Terra}} \\
\midrule

I'm done \#bipolar \#bipolarclub113
& L4 & L4 & L5 & L4 & L5 & L5 \\

I'm done \#bipolar \#bipolarclub113.1
& L4 & L4 & L5 & L4 & L5 & L5 \\

Dark too dark falling falling fading.....gone
& L5 & L5 & L5 & L5 & L5 & L5 \\

My current mood is.....probably not safe enough to tweet about11
& L2 & L2 & L5 & L2 & L4 & L4 \\

Fuckin ideation won't go away HELP @BipolarClubRx \#bipolar111
& L3 & L3 & L5 & L3 & L3 & L5 \\

It's useless I can't anymore BYE1
& L5 & L5 & L5 & L5 & L5 & L5 \\

\bottomrule
\end{tabular}
\parbox{\textwidth}{
\vspace{2pt}
\textit{Note:} All tweets in this table received L5 as the SME-derived reference label. L1--L7 represent the seven annotation categories, with L5 corresponding to ``Acute Crisis with Suicidal Ideation.''
}
\end{table*}

\textbf{LLM Prediction of Acute Crisis Cases.} Table~\ref{tab:llm-predictions-L5} shows how the six selected LLMs classified tweets assigned the highest-severity label, L5, by the human-derived reference label. Qwen2.5-7B and GPT-5.6-Terra appear more sensitive in assigning the highest-severity label, with Qwen assigning L5 to all six examples and Terra assigning L5 to five. In contrast, Llama3.1-8B, Hermes3-8B, and GPT-5.4-mini assigned L5 only for Examples 3 and 6. These two examples received unanimous L5 predictions, suggesting that models consistently recognize cases where suicidal intent is explicit rather than implicit. Example 4 appears particularly challenging, as Qwen2.5-7B was the only model to assign L5. Finally, the two nearly identical versions of Example 1 received identical predictions across all six models, indicating consistency in model behavior for similar inputs.

\section*{LLM Behavior Under In-Context Demonstration Perturbations}

Explainable AI (XAI) is particularly important in mental health applications, where automated predictions can influence the interpretation of sensitive psychological states and crisis-related content. Beyond predictive performance, understanding why a model produces a particular classification is important for transparency, trust, and responsible use in high-stakes settings. In the context of LLM-based annotation, examining how models respond to changes in in-context demonstrations can provide insight into the factors that influence their decisions. To further examine model sensitivity to few-shot examples, we introduced three additional prompt variants that modify the demonstrations by flipping their labels or adding complementary explanations, following the contrastive demonstration methodology of Liu et al. \cite{liu2023towards}.

\begin{table*}[tb]
\caption{Few-Shot Demonstrations Used in Prompts I and J}
\label{tab:prompt-ij-demonstrations}
\centering
\begin{tabular}{p{7cm}p{2cm}p{2.8cm}p{2.8cm}}
\toprule
Demonstration Tweet &
Reference Label &
Random Label-Flipped &
Targeted Label-Flipped \\
\midrule

Happy birthday to the beloved king of grunge \#[PERSON] &
L1 &
L6 &
L1 \\

I can't make it through this one guys I'm sorry &
L2 &
L1 &
L1 \\

[PERSON] has died rot in pieces you racist sack of crap &
L1 &
L1 &
L1 \\

\bottomrule
\end{tabular}
\end{table*}

\textbf{Prompt I -- Random Label-Flipped.} Uses the same three few-shot examples (see Table \ref{tab:prompt-ij-demonstrations}) as Prompt E but deliberately reassigns the demonstration labels to incorrect, plausible alternatives. This variant examines how dependent models are on the correctness of in-context demonstrations and measures the extent to which misleading but plausible labels affect classification performance.

\textbf{Prompt J -- Targeted Label-Flipped.} Uses the same three few-shot examples (see Table \ref{tab:prompt-ij-demonstrations}) as Prompt E but changes only the label of the crisis-related example to ``1-No Concern,'' while keeping the labels of the other two examples unchanged. The two correctly labeled examples serve as controls, allowing us to examine whether a single misleading demonstration, specifically one that contradicts the expected crisis interpretation, can influence model predictions. This provides a targeted test of model robustness to a misleading in-context demonstration.

\textbf{Prompt K -- Explanatory Demonstrations.} Retains the three tweets and their correct labels from Prompt E but adds a brief explanation describing why each example was assigned its corresponding label. Rather than asking the model to generate reasoning for the test tweet, this variant provides the reasoning as part of the demonstrations themselves. It therefore examines whether explanatory few-shot examples provide additional guidance beyond tweet-label pairs and separately evaluates the contribution of demonstration-level explanations. Across the evaluated models, Prompt K produced a Micro F1 closest to the Prompt E baseline on average, although the effect varied across models. Some models improved, whereas others declined, suggesting that adding explanations to demonstrations does not consistently improve annotation performance.

\textbf{Sensitivity to in-context demonstrations.} Models differed substantially in their sensitivity to corrupted demonstrations. The label-flipping experiments showed that many models experienced a decline in performance when demonstration labels were altered, whereas some models were comparatively insensitive to the manipulation. This variation suggests that models do not use in-context demonstrations uniformly: some appear to treat the demonstrations as a strong decision signal, whereas others may rely more heavily on pretrained knowledge and the category definitions.

\textbf{Complementary explanations do not uniformly help.} Adding explanations to the demonstrations did not consistently improve classification performance. Although Prompt K produced a Micro-F1 close to the Prompt E baseline on average, individual models showed different responses, with some improving and others declining. This suggests that the usefulness of explanatory demonstrations is model-dependent rather than providing a consistent performance benefit.

\textbf{Model-specific error patterns.} The class-level analysis revealed distinct error patterns across models, with some models consistently confusing particular neighboring severity categories. These patterns suggest that models differ not only in overall performance but also in how they interpret and separate crisis-severity boundaries.

\textbf{Reasoning and classification are not necessarily aligned.} The reasoning analysis suggests that generating an explanation does not guarantee a more accurate classification. Models may provide plausible reasoning while still exhibiting systematic errors on particular severity categories, indicating that the presence of an explanation alone is insufficient to establish reliable decision-making.

\textbf{Highest-severity crisis detection exposes a critical behavioral difference.} Aggregate annotation performance did not necessarily correspond to reliable detection of the highest-severity crisis category. The class-specific analysis showed that some models with relatively strong overall performance failed to identify the rare highest-severity cases, while other models with lower overall performance successfully identified them.

\section*{Conclusion}
This study evaluated LLMs for crisis-severity annotation of veteran-related social-media content against annotations from eight Veteran SMEs. Human reviewers showed low categorical agreement, while ordinal agreement was higher, indicating that disagreements often occurred between neighboring severity levels. LLMs showed substantially higher inter-model consistency, but greater consistency did not necessarily mean better alignment with human judgments.

Proprietary models generally achieved higher Micro F1 and greater stability than open-source models, while prompt design influenced model performance. LLMs also struggled with ambiguous and implicit crisis signals, particularly rare high-severity cases. In-context demonstration experiments further showed that models varied in their sensitivity to misleading examples, and explanatory demonstrations did not consistently improve performance. These findings concern agreement with an SME-derived annotation framework for historical social-media posts and do not establish clinical diagnostic accuracy. 

Overall, LLMs show promise as scalable complements to human annotation, but should not replace human expertise in safety-critical mental-health applications.

\section*{Acknowledgment}
This study was partly supported by the Northwestern Mutual Data Science Institute. The content is solely the responsibility of the authors and does not necessarily represent the views of the funder.

\bibliographystyle{IEEEtran}
\bibliography{Ref}
\vspace{12pt}

\end{document}